\documentclass{article} 
\usepackage{iclr2021_conference,times}

\usepackage{amsmath,amsfonts,bm}

\def\eqref#1{equation~\ref{#1}}

\def\1{\bm{1}}

\DeclareMathAlphabet{\mathsfit}{\encodingdefault}{\sfdefault}{m}{sl}
\SetMathAlphabet{\mathsfit}{bold}{\encodingdefault}{\sfdefault}{bx}{n}

\usepackage{hyperref}
\usepackage{url}
\usepackage{csquotes}

\title{Strengthening Subcommunities:\\ Towards Sustainable Growth in AI Research}

\author {
    Andi Peng, \textsuperscript{\rm 1,2\thanks{Equal contribution.}}
    Jessica Zosa Forde \textsuperscript{\rm 3*}
    Yonadav Shavit \textsuperscript{\rm 2,4*}
    Jonathan Frankle \textsuperscript{\rm 1,5*} \\
    \textsuperscript{\rm 1} MIT, \textsuperscript{\rm 2} Schmidt Futures, \textsuperscript{\rm 3} Brown, \textsuperscript{\rm 4} Harvard, \textsuperscript{\rm 5} MosaicML
}

\begin{document}

\maketitle

\begin{abstract}
AI’s rapid growth has been felt acutely by scholarly venues, leading to growing pains within the peer review process.
These challenges largely center on the inability of specific subareas to identify and evaluate work that is appropriate according to criteria relevant to each subcommunity as determined by stakeholders of that subarea.
We set forth a proposal that re-focuses efforts within these subcommunities through a decentralization of the reviewing and publication process. Through this re-centering effort, we hope to encourage each subarea to confront the issues specific to their process of academic publication and incentivization.
This model has historically been successful for several subcommunities in AI, and we highlight those instances as examples for how the broader field can continue to evolve despite its continually growing size.
\end{abstract}

\section{Challenges in the existing system}

\subsection{Exponential Field Growth}
AI’s rapid growth has been felt acutely by scholarly venues, particularly large field-wise conferences. A 2018 study noted that the number of submissions increased by “47\% for ICML, by 50\% for NeurIPS, and by almost 100\% for ICLR”~\citep{Sculley2018-qe}. More recent estimates by NeurIPS in 2020 put their year over year growth rate at a similarly high 40\% \citep{Neural_Information_Processing_Systems_Conference2020-kx}. This rapid growth has led to growing pains within the peer review process. While conferences in the past would use recommendations from meta-reviewers to generate new reviewer invitations, conferences in most recent years have resorted to soliciting reviewers over social media to fill the reviewer gap. In 2021, NeurIPS asked reviewers to offer to review (positively bid) 30-40 papers out of an estimated 12,000 submissions. These bids resulted in reviewers matched with 6-7 papers each, with instances of AAAI reviewers being asked to review as many as 10 papers.

\subsection{Low Reviewer Quality}
Relieving reviewer burden has resulted in an increasing number of reviewers with limited reviewing experience. Lack of experience degrades reviewer quality.  In 2019, the shortest reviews for ICLR 2020 were as little as 17 words \citep{Sun2019-zf}. These lackluster reviews recommended rejection, yet often the reviewer admitted a lack of experience in the subject. Inexperienced reviewers also often express biased reviews within the reviewing process; researchers have noted that inexperienced reviewers tend to more easily reject conference re-submissions, yet one could argue these re-submissions are, on average, higher quality than first submissions \citep{Stelmakh2021-sb}. 

\subsection{Reviewer Inconsistency}
This poor quality of reviews is compounded by poor calibration and inter-rater reliability among reviewer pools. A recent analysis of the 2014 NeurIPS reviewing process, in which the number of submissions to the conference are a fraction to today’s, notes that when papers were submitted to two separate pools of reviewers, 25\% of papers received differing recommendations among review committees for acceptance, while only 13\% received consistent recommendation towards acceptance \citep{Cortes2021-wu}. Moreover, the average rating given by the review committee is not correlated with the number of citations seven years later.

\subsection{Disproportionate Impact on Junior Researchers}
Poor reviews are experienced particularly harshly by junior researchers who are responsible for conducting experiments in the field yet have difficulty receiving quality feedback on their work. Evidence from the 2014 NeurIPS study suggests that revise and re-submit as a technique may not be a useful solution; among the 1,264 papers rejected from NeurIPS that year, only 34\% were eventually published within a peer-reviewed venue \citep{Cortes2021-wu}. While senior researchers also feel the pains of poor reviews, they are less subject to the whims of each submission since they often submit multiple works under a principal investigator (PI) role. 


\section{Idea: Re-Focus Within Subcommunities}
Rather than aggregating all academic dissemination in machine learning within concentrated mega-conferences, we propose that creating more specialized subcommunities for academic publication will be beneficial to solving many of the challenges listed above. We highlight the reasons here:

\subsubsection{Context-specific Reviewing Guidelines}

At present, all machine learning papers submitted to major conferences are subject to the same reviewing criteria. Reviewing forms are generic, asking the same questions no matter the style of work, opening the door for individual reviewers to bring their own perspectives on the nature of what represents a publishable contribution.
Yet, different subareas of work necessarily require different kinds of reviewing with that subarea's standards of publication.
For example, a paper investigating the training dynamics of neural networks should be evaluated according to its scientific rigor and insights, not its ability to improve the state-of-the-art accuracy on a particular task; for an architectural innovation, the opposite is true.
Many kinds of work value understanding without regard for real-world impact (e.g., theoretical work), whereas the value of other work hinges on whether it has made a difference in the real world (e.g., research on systems and efficiency).
Even simple recent attempts to update reviewing criteria (such as adding broader impact statements) have been fraught with difficulty because they are much more salient to specific areas than to the entire machine learning community. The present situation, which implicitly requires any criteria updates to be relevant to the entire community, detracts from our ability to implement subarea specific reform.

At the moment, it is impossible to evaluate the merits of research in a context-specific fashion.
By publishing research in smaller communities, it would become possible to narrowly tailor reviewing forms and standards for publication to the needs of specific areas.
Reviewers could receive more direction on how to evaluate papers and do so in a more uniform fashion.
Reviewer pools would be consistent within an area, allowing for informal shared values to emerge alongside formal criteria.

\subsubsection{Investment \textit{In} and Accountability \textit{To} Subcommunities}
Right now, the sheer size of the machine learning community necessitates researchers serving as reviewers for topics beyond their main areas of work.
Authors, fellow reviewers, and area chairs are anonymous names with whom they have never interacted before and will never interact again.
The reviewer has little investment in the outcome of the reviewing process.

By making communities smaller and focusing them on specific areas, researchers at all levels of seniority (from established figures to junior students) will have incentive to invest in ensuring the quality of the reviewing process and can more easily be held accountable for failures to do so.
Reviewers and reviewees will be peers, collaborators, and problem-specific interlocutors, not generic members of a large anonymized community.
Poor quality reviewing will be visible to intellectual colleagues with whom the reviewers will need to interact closely over the course of a career.
Researchers working on specific problems will have incentive to curate contributions to those literatures, as the result of the reviewing process will be acutely felt by the reviewers themselves.

\subsubsection{Experimentation of Novel Contributions}

Currently, new kinds of contributions are subject to old criteria that make it especially difficult for them to survive the reviewing process and form the basis of entirely new areas.
They must withstand stagnant reviewing forms and reviewers from other communities who will inevitably bring pre-conceived notions of novelty and significance that are incompatible with new perspectives.

By designating spaces for smaller subcommunities, it would become possible to experiment with new reviewing criteria designed to address the nuances of emerging contributions.
Machine learning for healthcare, AI fairness, systems, and mechanism design for social good subcommunities all began with smaller workshops within larger conferences. With each subsequent version of the workshop, researchers were able to build more and more robust subcommunities and define the parameters under which work would be evaluated and shared. These forms of research experimentation led to the formalization of new fields and the creation of new conferences.

Newly-formed conferences now have the flexibility to define their unique paper calls to encourage collaboration between machine learning researchers and relevant stakeholders in their fields. For example, the ML for healthcare conference, CHIL, provides tracks for \enquote{Applications and Practice} and \enquote{Policy: Impact and Society} built upon the shared goal of creating systems that can positively benefit individual and population-level health. The mechanism design conference, EEAMO, and the fairness and accountability conference, FAACT, similarly encourage pedagogically aligned work.


\section{How do we create more subcommunities?}

\subsection{(Idea 1) Prioritize more focused workshops in large conferences} 
According to NeurIPS and ICLR, \enquote{Good workshops have helped crystallize common problems, explicitly contrast competing frameworks, and clarify essential questions for a subfield or application area.}. Empowering workshop organizers to seek emerging common topic areas will allow for centralization of thought partners. Moreover, conferences can identify the successes from each workshop and incorporate areas of high growth into the topics listed within the call for papers. 

\emph{Recommendations}:
\begin{itemize}
    \item Create communication channels between past workshop chairs and current conference program chairs to identify growing topic areas based on successful workshops. Then, build into the conference call for papers explicit encouragement of growing areas.
    \item Make explicit the goal of workshops to incubate and then integrate growing subfields into machine learning (i.e. incentivize follow-up gatherings and collaborations).
    \item Solicit and utilize retrospectives from workshop organizers as a means of finding ways to improve in the following year, connecting the thread of past and future subarea work.
\end{itemize}

\subsection{Idea 2) Create focused conference spinoffs dedicated to serving a particular subcommunity}
In the past few years, we have seen a slow but steady rise of spinoff conferences dedicated to focusing academic publication within identified subareas. For example, both the ACM Conference on Fairness, Accountability, and Transparency (FAccT) and Conference on Robot Learning (CoRL) began with the intention to bring across diverse researchers within in specialized spaces. Over time, we have seen how submitted work within each subarea has shifted towards these venues being influential venues for dissemination, and the rise of conference submission and attendance as a result. We can follow this model to help subcommunities currently neglected by the larger community.

\emph{Recommendations}:
\begin{itemize}
    \item Identify subareas in machine learning that lack a dedicated, focused conference venue.
    \item Work with leading academics in those subareas to scope the feasibility of a conference spinoff. Then, engage these influential stakeholders in the conference creation process.
    \item Solicit funding and institutional support (look to CoRL and FAccT as working models) in creating a playbook to facilitate parallel workstreams for different subareas.
\end{itemize}

\subsection{Idea 3) Federate the mega conferences}
Major ML conferences already constitute multiple colocated conferences (e.g. large language modeling, DL theory, graph NNs).
However, these functional communities are kept informal, and thus cannot benefit from self-governance mechanisms: the ability to curate an in-community reviewing pool, to collectively define and enforce norms for the community, and to build and benefit from a more well-defined brand.
One possible solution is to ``federate'' existing large conferences, either by creating more formal in-conference ``tracks'', or by simply spinning off into many colocated smaller conferences which have logistics handled by a central body (as is the case in FCRC~\cite{fcrchomepage}).

The NeurIPS dataset track is an useful example of a federation-like model in practice.  In 2021, NeurIPS created a new track focused on datasets and benchmarks, which led to the publication of papers that have been historically overlooked. The chairs of the track noted that prior conferences had published \enquote{very few (less than 5) accepted papers per year focus on proposing new datasets, and only about 10 focus on systemic benchmarking of algorithms across a wide range of datasets} \citep{Vanschoren2021-hm}. As a parallel track separate from the main conference track, datasets and benchmarks encouraged deeper discussions of evaluation in ML. Moreover, it shifted the focus of reviewers from typical goal of beating a particular benchmark to reconsidering the role of a benchmark in furthering the field. Because NeurIPS is able to accommodate novel tracks as part of the main conference, we believe that it can continue this with growing research subcommunities.

\emph{Recommendations}:
\begin{itemize}
    \item Conference organizers should allow subcommunities to apply for and create formal subarea ``tracks'' (likely initially as outgrowths of successful workshops).
    \item Conference organizers should assist these tracks by providing tools (structural overhead, PR, etc.) for them to help self-govern, advertise, and solicit work.
\end{itemize}

\section{Limitations}

\subsection{Stratification}
Interactions between research communities have led researchers to adopt relevant ideas from other subareas. Successes in deep learning for computer vision eventually spread to NLP and reinforcement learning. Similarly, transformers, which were developed in NLP, are more recently being adopted in RL and CV. We are supportive of cross-pollination and do not wish to hinder these trends. It is possible that creating venues which occur at separate times and places as a large multi-topic conference would inhibit researchers in other fields to be exposed to outside work.

\subsection{Subcommunity accountability}
Best practices and calls for accountability are also important between subcommunities. For example, emphasis on reproducible experimental practice was osmosed into RL and the field's connections to machine learning as a whole led to reputability challenges and consistent code submission. Separating subfields could lead researchers to dismiss critiques from outside their area of expertise.

\subsection{Logistical Overhead}
Creating additional conferences for subcommunities may increase logistical burden.  We are optimistic, however, that these additional costs are offset by the ability of program chairs to divide and conquer reviewing and organizing load, perhaps into separate conferences and reviewing times.

\section{Conclusion}
We have presented a theory of change for alleviating many of the growing pains felt within ML publishing. Our proposal centers on re-focusing subares within specialized communities that can better serve their needs. In this way, we hope to return academic publishing back to the key stakeholders of each subarea, allowing for our work to grow (and be reviewed) sustainably.






\bibliography{iclr2021_conference}
\bibliographystyle{iclr2021_conference}

\end{document}